\documentclass[preprint,12pt]{elsarticle}

\usepackage[utf8]{inputenc} 
\usepackage[T1]{fontenc}    
\usepackage{url}            
\usepackage{booktabs}       
\usepackage{amsfonts}       
\usepackage{nicefrac}       
\usepackage{microtype}      
\usepackage{lipsum}
\usepackage{graphicx}
\usepackage{hyperref}
\hypersetup{
    colorlinks=true,
    linkcolor=blue,
    filecolor=blue,      
    urlcolor=blue,
    citecolor=blue,
    pdftitle={Overleaf Example},
    pdfpagemode=FullScreen,
}

\usepackage{fancyhdr}
\usepackage[symbol]{footmisc}
\usepackage[font=small, labelfont=bf]{caption} 
\usepackage{subcaption}
\usepackage{float}
\usepackage{adjustbox}
\usepackage{amsmath}
\usepackage{multirow}
\usepackage{adjustbox}
\usepackage{multicol}
\usepackage{lscape}
\usepackage{longtable}
\usepackage{lipsum}
\usepackage[colorinlistoftodos,disable]{todonotes}
\usepackage{diagbox}
\usepackage{pdflscape}
\usepackage{setspace}
\journal{Knowledge-Based Systems}

\begin{document}

\begin{frontmatter}

\title{ElectroCardioGuard: Preventing Patient Misidentification in Electrocardiogram Databases through Neural Networks}

\author[inst1]{Michal Seják (sejakm@kiv.zcu.cz)}
\author[inst1]{Jakub Sido}
\author[inst2]{David Žahour}

\affiliation[inst1]{organization={Department of Computer Science and Engineering}}

\affiliation[inst2]{organization={Department of Cybernetics},
            addressline={Technická 8},
            city={Plzeň},
            postcode={32600},
            country={The Czech Republic}}

\begin{abstract}

\noindent Electrocardiograms (ECGs) are commonly used by cardiologists to detect heart-related pathological conditions. Reliable collections of ECGs are crucial for precise diagnosis. However, in clinical practice, the assignment of captured ECG recordings to incorrect patients can occur inadvertently. In collaboration with a clinical and research facility which recognized this challenge and reached out to us, we present a study that addresses this issue. In this work, we propose a small and efficient neural-network based model for determining whether two ECGs originate from the same patient. Our model demonstrates great generalization capabilities and achieves state-of-the-art performance in gallery-probe patient identification on PTB-XL while utilizing 760x fewer parameters. Furthermore, we present a technique leveraging our model for detection of recording-assignment mistakes, showcasing its applicability in a realistic scenario. Finally, we evaluate our model on a newly collected ECG dataset specifically curated for this study, and make it public for the research community.

\end{abstract}

\end{frontmatter}

\newcommand{\aA}{\textit{vec-avg}}
\newcommand{\aB}{\textit{disc-avg}}
\newcommand{\aC}{\textit{weighted-disc-avg}}
\newcommand{\aD}{\textit{weighted-consistency}}

\section{Introduction}


Accurate interpretation of electrocardiogram (ECG) recordings is crucial for achieving high diagnostic accuracy and minimizing the risk of errors during diagnosis establishment and consecutive treatment. Unfortunately, in administrative practice, instances occur where physicians capture ECG recordings and assign them to incorrect patients by mistake. This issue increases the risk of inaccurate diagnoses and can have serious implications for patients.

In recent years, medicine has witnessed significant advancements through the utilization of advanced technologies and automated systems to enhance diagnostic accuracy and treatment procedures. Artificial neural networks, as a technique of artificial intelligence, have gained increasing prominence in the healthcare sector and hold the potential for revolutionary changes.

It should be emphasized that hospitals often face financial constraints that prevent them from acquiring expensive technologies or hardware not directly used for healthcare. For this reason, it is essential to utilize computationally inexpensive models that are affordable and easy to implement in the hospital environment. These models should be efficient in the analysis of ECG recordings and could provide an alternative to more expensive technologies without compromising the quality of diagnosis and patient care. 


This work is motivated by a hospital seeking a solution to address the issue of errors that occur in their database of ECG recordings, which can have significant implications for patient diagnosis and subsequent treatment.

In this paper, we focus on the development and evaluation of an automated system for detection of misclassification of ECG recordings using a modern convolutional network architecture (CDIL-CNN \cite{cheng2023classification}) suitable for processing sequential data. Our objective is to minimize the number of patient and recording mix-ups prior to ECG evaluations, which can lead to inaccurate diagnoses and compromised patient care. We believe the system could significantly reduce the number of errors in administrative ECG evaluations and contribute to higher accuracy and reliability of diagnoses. 


This work makes significant contributions in several key areas:
\todo{Updated 'low computational requirements' to 'manageable computational requirements'}
I. The development of an exceptionally efficient system with manageable computational requirements allows for practical implementation in resource-constrained settings, specifically benefiting hospitals with limited computational resources. This aspect ensures the model's viability and effectiveness in real-world healthcare environments.
II. We achieved improved performance by significantly reducing the number of model parameters 760+ times in comparison with concurrent works. This enhancement increases the model's efficiency without compromising its accuracy, providing a more streamlined and effective solution for ECG analysis.
III. Additionally, we present an anonymized version of a real-world dataset specifically curated for this study. This contribution not only advances the field of ECG analysis but also provides a valuable resource for future research in this domain. The availability of this dataset fosters collaboration and facilitates further investigations into improving diagnostic accuracy and patient care.
IV. The work further simulates the deployment of the developed model under realistic conditions, replicating its application within a clinical setting. By mimicking real-world scenarios, this simulation provides insights into the model's performance and feasibility in practical use, offering valuable perspectives on its potential impact in improving ECG evaluations and supporting healthcare professionals in their decision-making process.

In summary, this work's main contributions include the utilization of an efficient model for resource-constrained settings, achieving improved performance through parameter reduction, the publication of an anonymized real dataset, and simulating real-world deployment. These advancements offer promising avenues for enhanced ECG analysis, opening doors to more accurate diagnoses and improved patient care.


\section{Related Work}
\label{sec:rel_work}

Early approaches to ECG-based patient identification rely on hand-crafted feature extraction methods (FEMs) completely \cite{biel2001ecg, saechia2005human}. Some recent works, while utilizing the representational power of neural networks or other function approximators, still utilize preliminary feature extraction methods \cite{karpagachelvi2010ecg, el_rahman2019biometric, patro2017effective, ko2019ecg, boumbarov2009ecg}. These methods include R-peak detection \cite{qin2017adaptive, kaur2019novel, rajani2021r, gupta2020r, laitala2020robust} and segmentation \cite{kim2020ecg}, measuring onsets and durations of different beat phases \cite{biel2001ecg, ko2019ecg, pathoumvanh2013ecg}, Kalman filter transformation \cite{ting2010ecg} or PCA transformation \cite{boumbarov2009ecg}.

FEMs are more resistant to noise in the original signal and therefore less prone to over-fitting. However, the choice and tuning of a specific FEM depends on prior expert domain knowledge and is therefore subjected to possible human error and bias. Moreover, transforming the input signal to a different domain using FEMs is likely to -- in addition to noise -- remove a part of the information relevant to the task we are trying to solve. Suffice to say, designing and optimizing FEMs automatically is difficult.

With the rise of deep-learning methods and availability of better hardware, researchers have applied deep neural networks to the task of of identifying patients using the ECG signal directly \cite{labati2019deep, jyotishi2020lstm, hammad2018multimodal, jyotishi2021ecg}. Such models not only require large amounts of data for training and are prone to over-fitting, but are also sensitive to noise, which commonly occurs in the ECG signal; wandering baseline, interference caused by power lines and other electrical devices, muscle contractions, motion artifacts and other high-frequency noise to name a few \cite{nayak2012filtering}. To address the presence of noise, researchers pre-process the signal \cite{li2017genetic, lyakhov2021system, tyagi2021intellectual, ji2019electrocardiogram, perez2016application} using 
various high-pass, low-pass or band-stop/pass filters implemented using Fourier transformations, \cite{singhal2020efficient} wavelet transformations \cite{lyakhov2021system, chen2015hardware}, or other techniques \cite{labati2019deep}.

Regarding neural network architectures, prior art consists mostly of biometric systems based on convolutional neural networks (CNNs) \cite{labati2019deep, deshmane2018ecg, hammad2021resnet, hammad2018multimodal, melzi2023ecg} and recurrent neural networks implemented using long-short term memory cells (LSTMs) \cite{kim2020ecg, jyotishi2020lstm, jyotishi2021ecg}. LSTM networks represent a natural solution to signal processing as they are designed specifically to handle long, sequential data \cite{hochreiter1997long}. However, data cannot be passed through LSTM networks in parallel due to their sequence-processing nature. That motivates the use of CNNs, whose convolutional filters can be parallelized on GPUs very efficiently. In the context of ECG biometrics, CNNs can be either applied to electrocardiogram images using two-dimensional filters \cite{hanilcci2019ecg, lee2022personal, labati2019deep} or directly to the input signals using one-dimensional filters \cite{hammad2021resnet}.

A noticeable feature of many previous works is the size of datasets commonly used for training and evaluation of patient identification systems. Such datasets include the MIT-BIH Normal Sinus Rhythm (NSRDB) and MIT-BIH Arrhythmia (MITDB)  \cite{moody2001impact, goldberger2000physiobank}, CYBHi \cite{da2014check}, ECG-ID \cite{lugovaya2005biometric} and PTB \cite{bousseljot1995nutzung}, which contain tens or hundreds of patients at most. Patient ECGs in small datasets are very unlikely to be representative and independent samples of the whole population either due to biases introduced during the selection of patients to monitor or due to specific features of the devices or processes used to collect ECG signals. Evaluation on such data therefore provides limited proof of a system or a model generalizing well beyond the study's scope. 

Furthermore, the patient identification task is generally phrased as a direct multi-class classification problem \cite{melzi2023ecg, labati2019deep, hammad2021resnet, jyotishi2020lstm, jyotishi2021ecg, ghazarian2022assessing}, which may be motivated by the presence of small patient databases. In other words, researchers build models that directly map ECG signals to patients using only their trained parameters. Researchers commonly achieve this by designing neural network architectures whose output is a soft-max layer with one "neuron" for each patient in their dataset. The downside of this approach is the fact that such model -- in its full form -- is incapable of generalization, because the set of patients it works with is already predetermined by its design. The implications are mostly practical: the model cannot be shared for or among different hospital environments, nor can it handle changes in the set of patients; in that case, it has to be fine-tuned again or retrained from scratch.

\todo{Added the following paragraph (In the context...)}
In the context of this work, the term "embedding" refers to a collection of numerical values that encapsulate the characteristics of a specific object, enabling it to be represented in a meaningful way. This concept can be thought of as both a tangible result – a set of numbers that capture the essence of the object – and a process, where data is transformed into these meaningful numerical representations. Specifically, an "ECG vector" signifies the outcome of embedding an ECG signal within a D-dimensional vector space. In simpler terms, an ECG vector is the representation of an ECG reading using a set of numbers that encode its distinctive features. The method employed to generate such ECG vectors is facilitated by an "embedding model," which is a type of model designed to create embeddings or these numerical representations for various objects. In general, the field concerned with creating embedding models is called representation learning. 

When attempting to generalize beyond the known set of patients, Jyotishi et al. \cite{jyotishi2021ecg} create these ECG vectors by discarding the soft-max layer after training their multi-class classification model, and instead think of the network's output at the last-but-one layer as an embedding of the ECG signal. To confirm the identity of a patient whose ECG readings are already known, they embed the new ECG in an ECG vector space (creating an ECG vector $p_a$) and compare it to the average of the patient's other ECG vectors $\bar{p_b} = \frac{1}{k} \cdot (p_{b1} + p_{b2} + ... + p_{bk})$ through cosine similarity. The decision whether the new ECG actually originates from the patient whom $p_{b1}$ to $p_{bk}$ belong to is obtained by comparing the result of the cosine similarity to some pre-determined threshold value \cite{jyotishi2021ecg}. Although this configuration appears satisfactory, it's uncertain whether cosine similarity can effectively compare ECG vectors of both familiar and unfamiliar patients, given the added complexity of the soft-max layer in transforming an ECG vector into a patient ID. Furthermore, limiting ourselves to the use of cosine similarity can cause our embedding model to "run out" of space on the hyper-sphere of ECG vectors to allocate to new patients. Even if the model was capable of perfectly embedding previously unseen patient ECGs, then for any threshold we choose, we can find a number of patients large enough for which the model starts failing to discriminate well between individuals. 

\todo{Added the following paragraph (Representation learning ...)}
Representation learning in general is a great approach to the problem we are trying to solve, namely not having to retrain our model when new patients are introduced. In recent years, representation learning saw noticeable progress in context of electrocardiography. The standard approach in representation learning is training representations using positive (similar) and negative (dissimilar) pairs, which are embedded close, \textit{resp.} far from one another, and thus the primary objective in representation learning is procuring these training pairs.

\todo{Added the following paragraph (SimCLR is a ...)}
SimCLR \cite{chen2020simple} is a self-supervised learning framework designed for learning meaningful visual representations from unlabeled data. The core idea behind SimCLR is to maximize the agreement between different augmentations of the same ECG recording while minimizing the agreement between augmentations of different ECG recordings. This way, the model learns to embed similar ECGs close to each other in the learned feature space. 3KG \cite{gopal20213kg} works in a similar way, but makes use of different data augmentation techniques, where the ECG is first transformed into a vectorcardiogram (VCG) using the Inverse Dowers transformation, randomly rotated and scaled, and converted back, while applying time-masking to the resulting ECG. Wav2Vec 2.0 \cite{baevski2020wav2vec} is a speech model with a convolutional encoder generating latent speech representations, a Transformer capturing sequence context, and a quantization module for discrete representation. Pre-training involves masked encoder outputs, which is again a data augmentation technique, while fine-tuning adapts it for speech recognition tasks using a linear projection and data augmentation for improved results. Lastly, CMSC \cite{kiyasseh2021clocs} stands for Contrastive Multi-Segment Coding, and it is one of the techniques investigated by Kiyasseh et al. for obtaining positive pairs from ECG recordings. It focuses on learning representations of adjacent non-overlapping temporal ECG recording segments. By capturing temporal invariances, CMSC demonstrates its potential in exploiting unlabeled physiological data. 

Oh et al. \cite{oh2022lead} have designed a model based on convolutional networks and Transformers \cite{vaswani2017attention} which utilizes a combinations of the above techniques. They have fine-tuned it for patient classification using the ArcFace loss \cite{deng2019arcface}, which indirectly trains the model's penultimate layer to produce valid ECG vectors in the context of patient identification. However, since their model is quite large, it may be difficult to deploy it in a hardware-constrained environment.

In summary, the issues we have discovered in related work are the following: multi-class patient classification, the use of large models, and evaluation on small datasets. Our study addresses these issues by:
\begin{itemize}
    \item designing a small model, suitable for deploying on low-end hardware, which decides whether two ECG recordings originate from the same individual
    \item training, and -- more importantly -- evaluating our model on large public datasets (\textit{cca.} 1000x larger than PTB and others)
\end{itemize}


\section{Method}

Instead of treating the patient identification task as a multi-class classification problem, we approach the task of patient identification indirectly. Instead of having a neural network classify ECG signals into patient classes, we created a model to decide whether two different ECG signals originate from the same patient or not. Such a model does not require us to know the amount of patients beforehand and the set of patients for identification can change over time without the need to retrain the model.

We then train our model in two phases. In the first phase, an embedding model learns to build the ECG vector space through metric learning \cite{kaya2019deep}, which is a technique of building vector representations of inputs based on a metric function, such as the Euclidean distance. In the second phase, the embedding model is augmented by a discriminator head and fine-tuned on ECG pairs. The discriminator head's inputs are two ECG vectors created by the embedding model, and its output is a probability estimate of whether those vectors belong to the same individual.

Finally, to verify that a specific patient indeed owns a new, previously unseen ECG signal, we utilize a database of patients, which are clusters of previously classified ECG vectors (embedding model outputs). First, we convert the signal to an ECG vector using the embedding model and then estimate the probability that the vector belongs to the selected cluster using the discriminator head. 

\begin{figure}
    \centering
    \begin{subfigure}[b]{\textwidth}
    \centering
    \includegraphics[width=\textwidth]{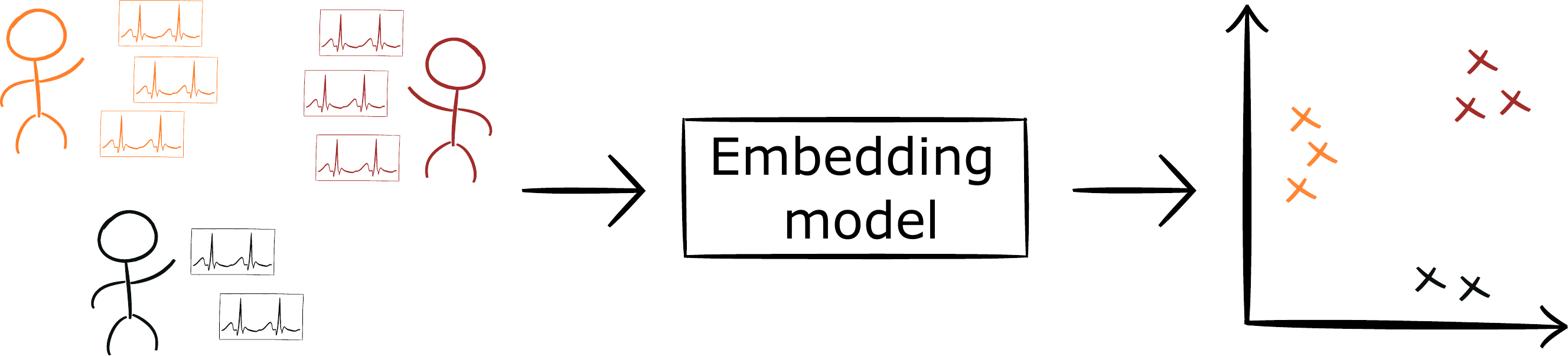}
    \caption{The first phase of training, which creates an ECG embedding model based on convolutional operations. The model maps ECG recordings to ECG vectors, thus describing an ECG vector space. In an ideal state, vectors (points) from the same patient are close to one another and far away from other patients -- ECG vectors from the same patient have the same color. This model is described in Section \ref{sec:architecture}.}
    \label{fig:overview-p1}
            \vspace*{7mm}
    \end{subfigure}
    \begin{subfigure}[b]{\textwidth}
    \centering
    \includegraphics[width=\textwidth]{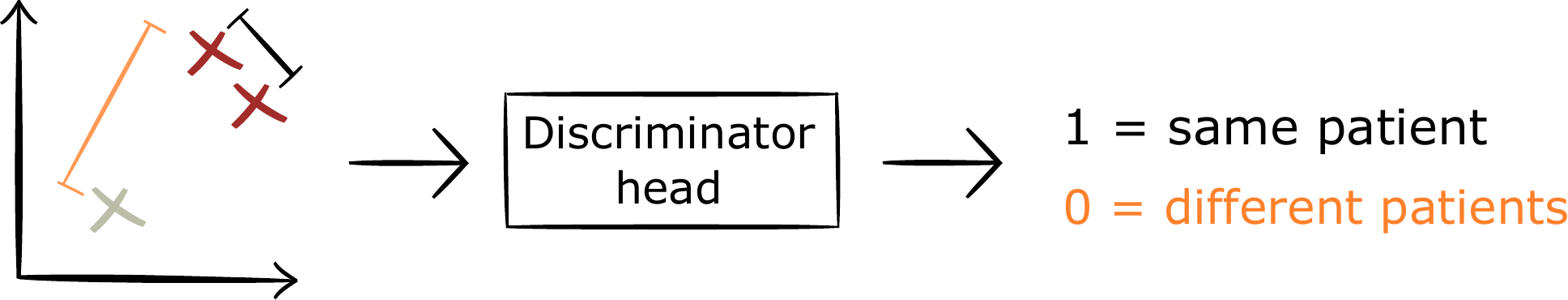}
    \caption{The second phase of training, which produces a small discriminator head. It compares two vectors from a pair using linear combinations of weighted distances between them and outputs the probability that the recordings, which were converted to these vectors by the embedding model, originate from the same patient. The discriminator learns to output 1 for positive pairs (same patient) and 0 for negative pairs (different patients). This model is described at the end of Section \ref{sec:architecture}.}
    \label{fig:overview-p2}
            \vspace*{7mm}
    \end{subfigure}
    \caption{The overview of our ECG-based patient identification method.}
    \label{fig:overview}
\end{figure}

\begin{figure}[h]
    \centering
    \includegraphics[width=\textwidth]{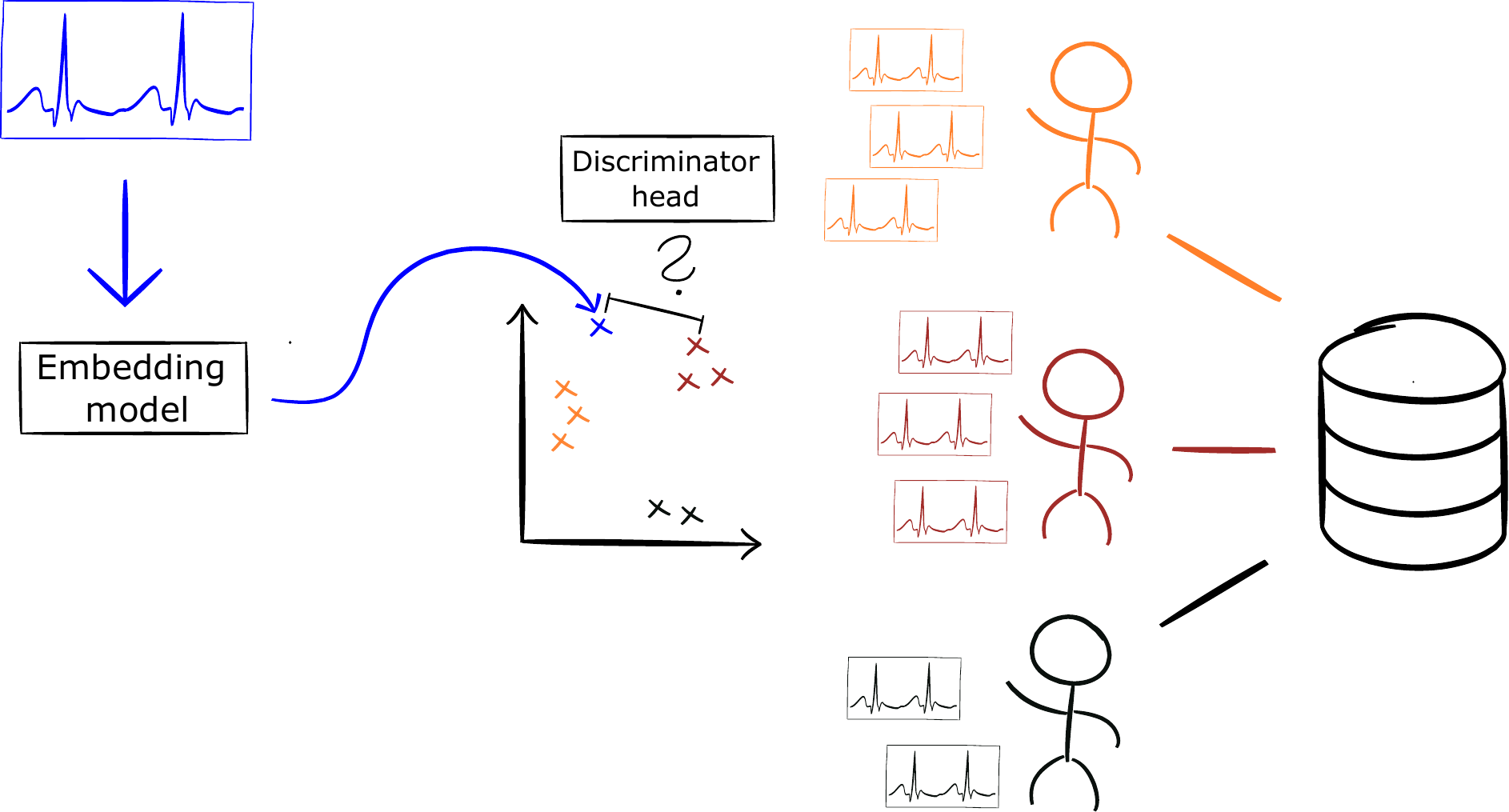}
    \caption{Schema of the intended deployment scenario. The hospital staff captures a new ECG recording (blue) and assigns it to a given patient (red). First, our embedding model converts the ECG signal to an ECG vector. Then, all ECG vectors belonging to the selected patient are loaded from a database. Finally, we use the discriminator head to calculate the likelihood that this is indeed the patient who the new ECG recording belongs to. The procedure is described in detail in Section \ref{sec:mistake-detection}, paragraph "Overseer simulation".}
    \label{fig:overview-p3}
\end{figure}

The overview of this whole process is captured by Figures \ref{fig:overview} and \ref{fig:overview-p3}. For the purpose of reproducibility, we publish our code-base as a GitHub repository at \href{https://github.com/CaptainTrojan/electrocardioguard}{https://github.com/CaptainTrojan/electrocardioguard} .

\subsection{Datasets}

Before delving into the specifics of our training and evaluation procedures, we would like to introduce the datasets used for this purpose.

Our study makes use of four distinct datasets, one of which connects us to the prior art, PTB \cite{bousseljot1995nutzung}. This dataset not only contains 12-lead ECG signals, but is also the largest and most relevant among all datasets mentioned in Section \ref{sec:rel_work}.

The remaining three datasets are CODE-15\% \cite{ribeiro2020automatic}, PTB-XL \cite{wagner2020ptb}, and a private collection of electrocardiogram data provided by the Institute for Clinical and Experimental Medicine (Prague, Czech Republic), which we refer to as "IKEM" in the context of our study. Table \ref{tab:datasets} shows their respective sizes. Notably, the newly created IKEM dataset is roughly four times larger than PTB-XL and contains more samples per patient on average, which is very useful for tasks involving intra-patient ECG comparison or matching. \todo{Added the following sentence (It consists of...)}
It consists of 12-lead recordings collected from patients which were examined by cardiology or diabetology sections at IKEM, sampled at 500 Hz for 10 seconds. 

\begin{table*}
    \centering
    \begin{tabular}{lllll}
        \toprule
        Title & \textnumero\ ECGs & \textnumero\ patients & Size & Train/dev/test split \\
        \midrule
        PTB & 549 & 290 & 69MB & 0/0/100 \\
        PTB-XL & 21799 & 18869 & 2.1GB & 0/50/50 \\
        IKEM & 98130 & 30290 & 6.3 GB & 0/50/50 \\
        CODE-15\% & 345106 & 233479 & 22GB & 70/10/20 \\
        \bottomrule
    \end{tabular}
    \caption{The datasets we use in our study. All datasets were separated into three non-overlapping sections for training, validation, and testing of our methods. The exact numbers can be retrieved using the \textit{dataset\_stats.py} script. 
    }
    \label{tab:datasets}
\end{table*}

\todo{Clarified the following paragraph (Each dataset...)}
Each dataset (including IKEM) contains raw ECG recordings paired with an anonymized unique patient ID, which defines clusters of ECG recordings belonging to the same patient. We store the recordings using 16-bit integers in HDF5 files with granularity 4.88 $\mu V$. We also remove redundant augmented leads and lead III, which can be calculated from leads I and II. These efforts to save storage space have resulted in size decrease of roughly 60\%, saving more than 50 gigabytes of data in total. When we input the signal to our model, we expand the reduced 8 leads back to the original 12, producing 12 voltage values for each time instance.

All leads, regardless of sampling frequency (400-500 Hz) or length (8-10 seconds) were bidirectionally truncated or padded with zeroes to 4096 voltage measurements over time. In the exceptional case of PTB, which contains approx. 30-100 second long signals sampled at 1000 Hz, we down-sampled the input signal to 500 Hz before applying bidirectional truncation. Hence, all our ECG recordings are matrices of size $(4096, 12)$. \todo{Added the following sentence (The patient...)} The patient ID, which is associated with each such matrix, is a single 32-bit unsigned integer.

\todo{Specified that dataset elements are tuples (ECG signal, patient ID) in the next sentence (The train/dev/test split...)}
The train/dev/test split is applied to the sequence of dataset elements, that is, tuples "ECG signal, patient ID", which means that -- in small numbers -- patients may be shared across splits. 


\subsection{Architecture}

\label{sec:architecture}

In the following sections, we describe the training and evaluation procedure captured by Figure \ref{fig:overview} in greater detail.

\paragraph{Pre-processing}

Before an electrocardiogram recording is input to our embedding model, it is pre-processed. As we have stated in Section \ref{sec:rel_work}, the raw ECG signal contains various types of noise. We have implemented a baseline wander (Figure \ref{fig:pp2}) and high-frequency noise removal filters (Figure \ref{fig:pp3}) using \textit{ptwt} (PyTorch Wavelets, \cite{Blanke2021}). Regarding power-line interference, both power-lines and ECG signals have varying frequencies across countries and data sources, so although the power-line interference noise may be present in our data as well, we have decided to omit this filter. Finally, we normalize the input signal to z-scores (Figure \ref{fig:pp4}) both in order to stabilize the gradient descent procedure and to calibrate the possibly different scales of electrocardiograms across different recording devices. Each of these transformations is applied to all leads separately and is a part of the embedding model, which makes the system resistant to noisy input.

\paragraph{Embedding model}

We have experimented with two different embedding model architectures: a one-dimensional residual convolutional network (1D-RN, Table \ref{tab:resnet}) \cite{he2016deep} and a circular dilated convolutional network (CDIL-CNN, Table \ref{tab:cdil}) \cite{cheng2023classification}, which is a novel architecture specifically designed for processing long sequences. These models pose an advantage over LSTM networks, Transformers \cite{vaswani2017attention} and state-space models \cite{smith2022simplified} as embedding models mostly thanks to their small size and short forward-pass duration. This enables hospitals to deploy our models on readily accessible hardware without the need to invest in high-end dedicated graphical processing units.

\begin{table}[!htb]
    \begin{subtable}{.5\linewidth}
      \centering
        \caption{}
        \begin{tabular}{ll}
            \toprule
            \textbf{1D-RN} & \\ 
            \textnumero\ parameters & 314M \\ 
            \midrule
            \textbf{Layer} & \textbf{Output size} \\
            \midrule
            & \\
            ECG signal & $(12, 4096)$ \\
            Pre-processing & $(12, 4096)$ \\
            Residual block 1 & $(128, 1024)$ \\
            Residual block 2 & $(196, 256)$ \\
            Residual block 3 & $(256, 64)$ \\
            Residual block 4 & $(320, 16)$ \\
            Flattening & $(5120, )$ \\
            3-layer perceptron & $(256, )$ \\
            \bottomrule
        \end{tabular}
        \label{tab:resnet}
    \end{subtable}%
    \begin{subtable}{.5\linewidth}
      \centering
        \caption{}
        \begin{tabular}{ll}
           \toprule
            \textbf{CDIL} & \\ 
            \textnumero\ parameters & 131K \\ 
            \midrule
            \textbf{Layer} & \textbf{Output size} \\
            \midrule
            ECG signal & $(12, 4096)$ \\
            Pre-processing & $(12, 4096)$ \\
            Initial block & $(32, 4096)$ \\
            7x Base block & $(32, 4096)$ \\
            Deformable block & $(32, 4096)$ \\
            Base block & $(32, 4096)$ \\
            Deformable block & $(256, 4096)$ \\
            Mean across time & $(256, )$ \\
            Dense layer & $(256, )$ \\
            \bottomrule
        \end{tabular}
        \label{tab:cdil}
    \end{subtable} 
    \caption{Summaries of evaluated embedding model structures. For further details regarding architecture or implementation of these models, see the original publications for ResNet \cite{he2016deep} and CDIL-CNN \cite{cheng2023classification}. The input ECG signal has 12 voltage values for 4096 samples through time. The output ECG vector dimension is 256. }
\end{table}


\paragraph{Metric learning}

To initialize the ECG vector space defined by the embedding model in a way that involves good separation of patients, we examine two deep metric learning approaches: \textit{triplet loss} \cite{hoffer2015deep, schroff2015facenet} and \textit{circle loss} \cite{sun2020circle}. Both loss functions accept three ECG vectors: anchor (A), positive (P), and negative sample (N), where the anchor and positive sample come from one patient and the negative from a different patient. Minimizing these losses corresponds to embedding the anchor and positive samples close to one another and far away from the negative sample. In other words, the goal is to maximize inter-patient distances and minimize intra-patient distances with respect to the individual ECG recordings. Figure \ref{fig:pca-embeddings} helps visualize what the end result might look like.


\paragraph{Discriminator head}

\begin{figure}[h]
    \centering
    \begin{subfigure}[b]{0.45\textwidth}
    \centering
    \includegraphics[width=\textwidth]{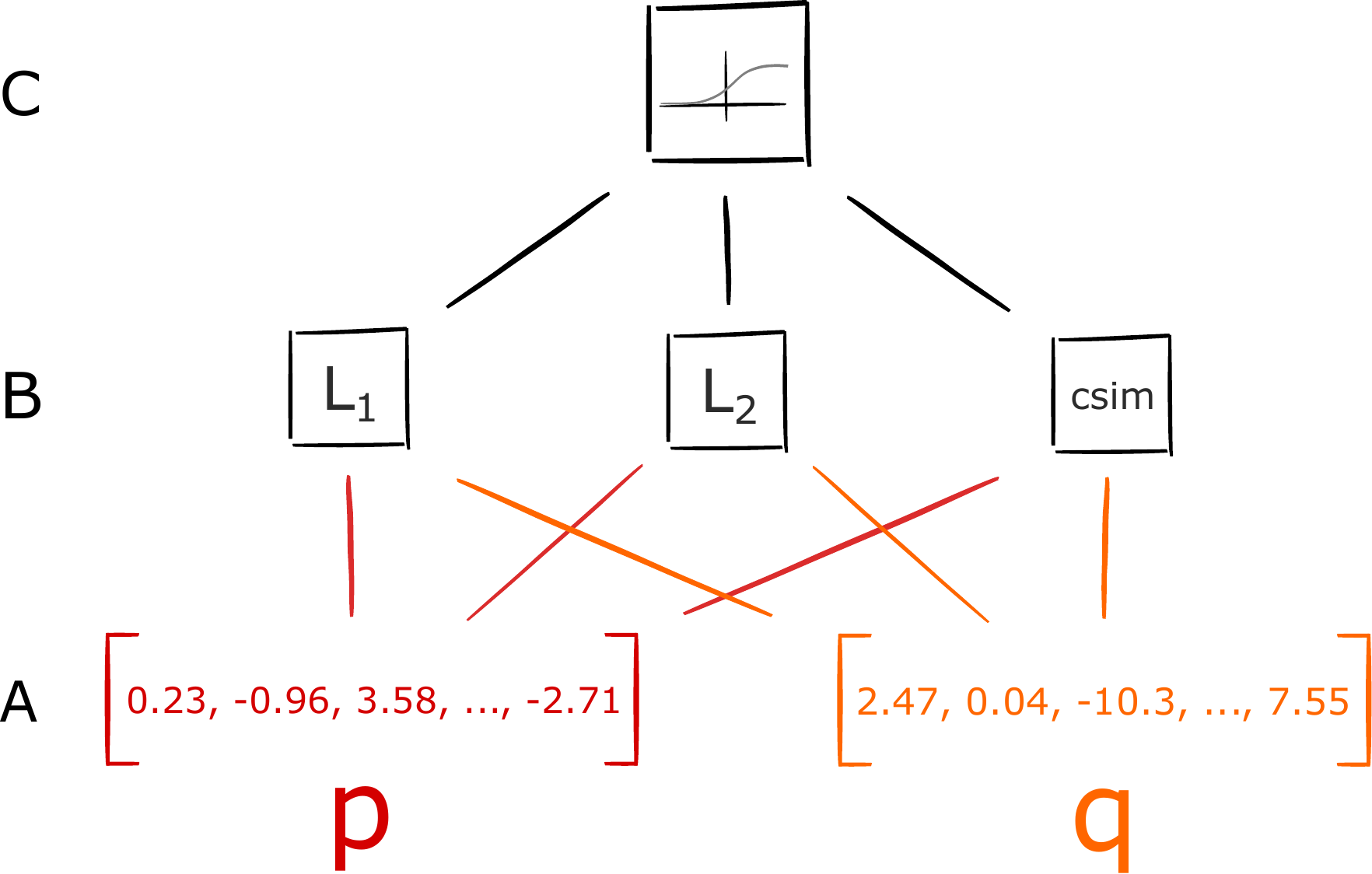}
    \caption{The discriminator head in detail. \textbf{A} -- Two input ECG vectors, $p$ and $q$. \textbf{B} -- Three distance metrics (Eq. \ref{eq:l1}, \ref{eq:l2} and \ref{eq:cos}) calculated between $p$ and $q$. \textbf{C} -- The results are linearly combined and squashed using the sigmoid activation. }
    \label{fig:discriminator-detail}
    \end{subfigure}
    \hfill
    \begin{subfigure}[b]{0.45\textwidth}
    \centering
    \includegraphics[width=\textwidth]{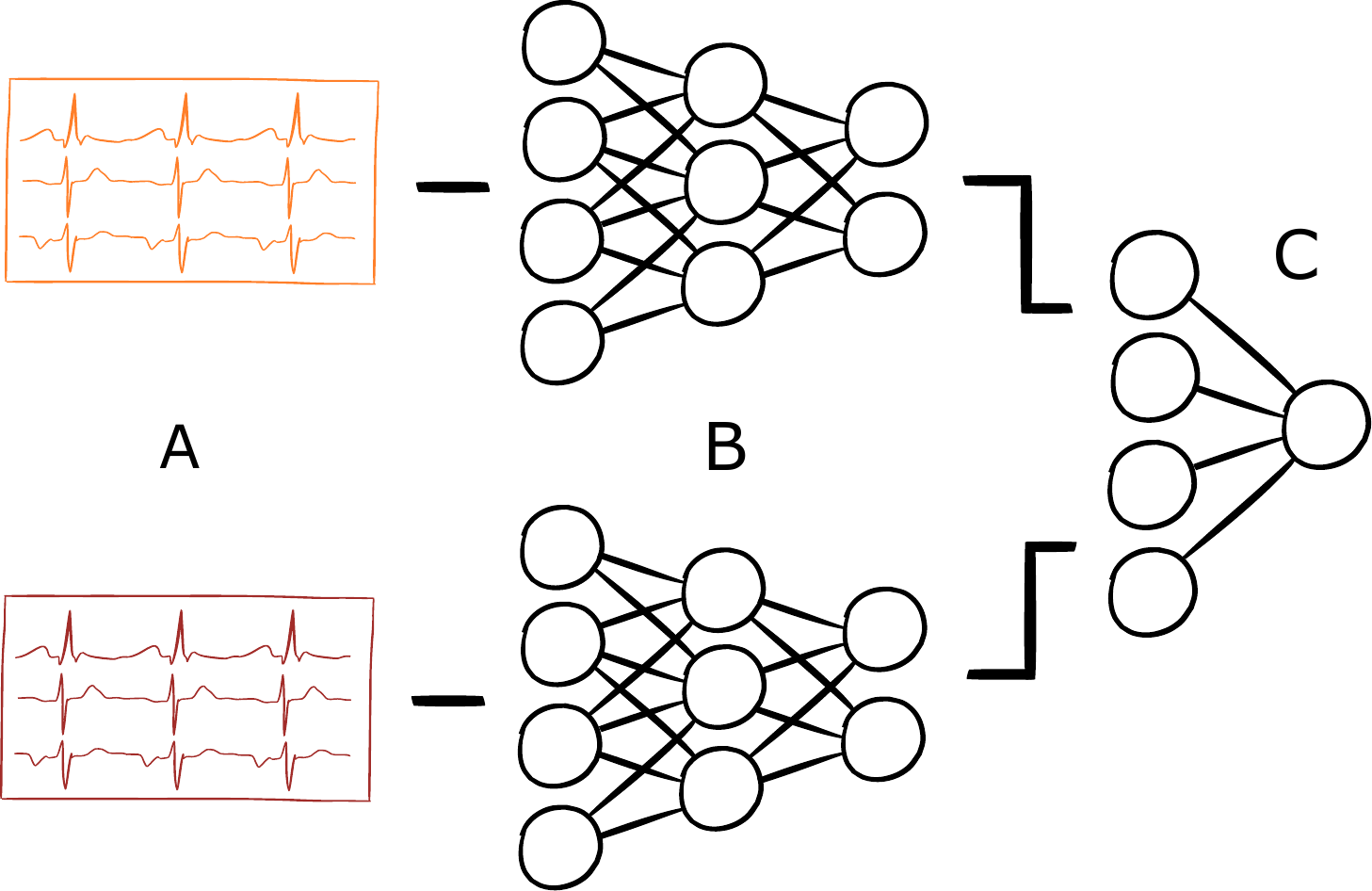}
    \caption{The Siamese model. \textbf{A} -- An input pair of ECG signals. \textbf{B} -- Duplicated embedding model converts both pair members to ECG vectors independently. \textbf{C} -- The discriminator head, which decides whether the input pair is positive (1) or negative (0). }
    \label{fig:siamese-model}
    \end{subfigure}
    \hfill
    \caption{The discriminator head and its role in the Siamese model, which is trained during the second phase.}
    \label{fig:discriminator-arch}
\end{figure}

After we finish training the embedding model, then, in order to utilize its ECG vector space initialization, we connect it to a small discriminator \textit{head} (a neural network whose inputs are the outputs of another neural network, see Figure \ref{fig:discriminator-detail}). The joint model resembles a Siamese setting (see Figure \ref{fig:siamese-model}), where the embedding model creates ECG vectors and the discriminator head decides whether they originate from the same patient.

\begin{align}
    \label{eq:l1}
    l_1(p, q) & = \sum_{i=1}^{N} |p_i - q_i| \\
    \label{eq:l2}
    l_2(p, q) & = \sum_{i=1}^{N} (p_i - q_i)^2 \\
    \label{eq:cos}
    csim(p, q) & =  \sum_{i=1}^{N} \frac{p_i \cdot q_i}{\sqrt{\sum\limits_{j=1}^{N}{p_j^2}} \cdot \sqrt{\sum\limits_{j=1}^{N}{q_j^2}}}
\end{align}

More specifically, the discriminator head's prediction is based on a linear combination of the distances between two input ECG vectors (Equations \ref{eq:l1}, \ref{eq:l2} and \ref{eq:cos}) that the embedding model creates. In order to let the network adapt to the embedding space's shape, we experiment with swapping the sum in the above formulas for a weighted sum instead whilst training those weights by gradient descent. The result is passed through a sigmoid layer in order to truncate the output to the range $(0,1)$, which is interpreted as probability. 

The Siamese model is trained on \textit{pairs} of ECG signals and learns to predict which pairs belong to the same patient and which belong to different patients, which is a binary classification problem. This model can be trained either end-to-end or with a frozen embedding model (its weights are never updated), so we experiment with both variants.

\subsection{Evaluation}
\label{sec:mistake-detection}

Finally, we evaluate our model on two tasks relevant to patient identification. The first task called gallery-probe matching is commonly used for this purpose in the context of identification based on similarity of representations \cite{cheng2011custom, hossain2010clothing, gross2006model, shen2018person}, including electrocardiogram-based patient identification \cite{oh2022lead}. Second, in order to evaluate the applicability of our model for ECG misclassification detection, we have designed a task we call the \textit{overseer simulation}. Both of these tasks are described in this section and our GitHub repository contains a stand-alone script \verb|pt_evaluate.py| that can be used for evaluating future models under the same conditions.

\paragraph{Gallery-probe matching}

The main objective of this task is to analyze the similarity of ECG vectors obtained from the same patient. To achieve this, we select a sample of N patients from a dataset, ensuring that each patient has at least two different ECG recordings, denoted as A and B. We divide these recordings into two distinct sets: the \textit{gallery} set, which contains all the A recordings, and the \textit{probe} set, which contains all the B recordings. Each set contains exactly one recording per patient. Our aim is to identify, for each probe element, the most similar gallery element, with the desired outcome of both elements belonging to the same patient. We measure the fraction of correctly matched pairs and report it as \textit{accuracy}. 

Here, the embedding model is first used for converting all $2N$ ECG recordings to ECG vectors, and the discriminator head $f_\theta(u, v)$ is used to calculate the similarity between members of all $N^2$ pairs.


\paragraph{Overseer simulation}

In the context of this task, we maintain a database of ECG vectors where each belongs to a specific patient from a dataset. This database is initialized by inserting at least one ECG vector (created by the embedding model) for N random patients from a dataset. Then, similarly to the previous task, we select another K random ECG recordings in total from some of those N patients, and aim to assign them correctly to their owners. The differences between this setup and the setup in gallery-probe matching are that both the database (gallery) and the probe set can contain multiple ECG recordings from one patient, and that both sets can have different size: database has at least N recordings, whereas the probe set has exactly K recordings.

Since we attempt to approximate the real use-case of our model, we simulate a hospital staff member whose task is to classify the probe ECG vectors instead of using the model directly. And to replicate the errors observed in practice, our simulated hospital staff member makes a \textit{mistake} (random patient chosen instead of the correct one) with some small probability $p$. Our model's role in this scenario is that of an \textit{overseer}, whose task is to detect these mistakes. Furthermore, as the elements from the probe set are classified, they are \textit{inserted} into the database, influencing the overseer's future decisions. If the staff makes a mistake and the overseer does not detect it, the database becomes corrupted. 

In summary, at each step of the overseer simulation, we are provided with an ECG vector and a patient selected by the staff, and our goal is to decide whether it is likely that this vector truly originates from the selected patient.

Our notation for the $j$-th ECG vector of patient number $i$ is $p_{ij}$. One patient $P_i$, then, is a set of vectors $p_{i1},\ p_{i2},\ ...,\ p_{in}$, where $n$ is the number of vectors originating from this patient. The vector being classified is called $v$, which is the output of an embedding model as seen in Figure \ref{fig:overview-p3}. We aim to calculate the measure of likelihood of $v$ originating from the patient $P$ selected by the staff, called $l(v, P)$, and if this value is smaller than some likelihood \textit{threshold}, we claim that the overseer has detected a mistake.

This decision should be based on the knowledge learned by the discriminator head $f_\theta(u, v)$. Nevertheless, we can only use the discriminator head when comparing $v$ to a single vector $u$, as in the gallery-probe matching task, but comparing $v$ to a \textit{set} of vectors $\{p_{ij}\}$ is not trivial. Therefore, we experiment with several different approaches of using the discriminator to calculate the likelihood, which are described in the remained of this section.




\subsubsection{\aA: Average of vectors}
An initial approach to calculating $l(v, P_i)$ is to replicate the approach by Jyotishi et al. \cite{jyotishi2021ecg}, which is to take an average of each vector $p_{ij}$ and use it as a single representative of $P_i$, as shown in Equation \ref{eq:approach_1}.

\begin{equation}
    \begin{aligned}
        \label{eq:approach_1}
        l(v, P_i) & = f_\theta (v, \bar{p_i}), \\ 
        \bar{p_i} & = \frac{1}{n} \sum_{j=1}^n p_{ij}
    \end{aligned}
\end{equation}

Such approach is however vulnerable to outlier vectors $p_{ij}$. Consider the case where an ECG signal measurement $p_{ik}$ is projected very far away from the general proximity of $P_i$'s other vectors: taking an average may now represent a completely different patient.

\subsubsection{\aB: Average of discriminator outputs}
Another simple approach is to take the average of the individual discriminator outputs between $v$ and every other $p_{ij}$, as shown in Equation \ref{eq:approach_2}.

\begin{equation}
    \begin{aligned}
        \label{eq:approach_2}
        l(v, P_i) & = \frac{1}{n} \cdot \sum_{j=1}^{n} f_\theta (v, p_{ij})
    \end{aligned}
\end{equation}

This approach is more resistant to outliers, as each vector is processed individually and contributes to the overall likelihood independently of others. However, it does not eliminate the threat of outlier noise completely, as distant vectors still have a large impact on the value of $l$. 

\subsubsection{\aC: Average of discriminator outputs with quality weighting}

To amend this issue, we can instead estimate the \textit{quality} of each $p_{ij}$ as a representative of $P_i$ and weigh the discriminator outputs by this quality value; see Equation \ref{eq:approach_3}. 

\begin{equation}
    \begin{aligned}
        \label{eq:approach_3}
        l(v, P_i) & = \frac{1}{\sum\limits_{j=0}^{n} q_j} \cdot \sum_{j=1}^{n} f_\theta (v, p_{ij}) * q_j, \\
        q_j & = q(p_{ij}, P_i) = \sum_{\substack{k = 1 \\ k \neq j}}^{n} f_\theta (p_{ij}, p_{ik})
    \end{aligned}
\end{equation}

This method guarantees that ECG vectors that the discriminator considers as poor representatives of the patient group $P_i$ will have a smaller impact on the general likelihood measure $l(v, P_i)$. Notice that $q_j$ does not have to be normalized itself, because $l(v, P_i)$ is normalized by the sum of $q_j$ regardless. Note that the difference between the previous and current approach is not recognizable until the cluster size is at least 3.

A possible downside of this approach is perhaps the fact that it fails to utilize the knowledge that some clusters are coherent (according to the discriminator) and some are not. Although the weighting itself allocates more impact to representative vectors $p_{ij}$, it does not capture the fact that such a weighting was even necessary.

\subsubsection{\aD: Previous approach with cluster consistency weighting}

The main idea behind \aD\ is that we calculate an estimate of a consistency measure $c_i$ of $P_i$ and should think of the consistency as an indicator of how much can we rely on the information provided by the intra-cluster and cluster-to-$v$ discriminator outputs. The exact formulation is captured by Equation \ref{eq:approach_4}, where $l'$ is Equation \ref{eq:approach_3}.

\begin{equation}
    \begin{aligned}
        \label{eq:approach_4}
        l(v, P_i) & = c_i \cdot l'(v, P_i), \\
        c_i & = \frac{1}{n(n -1)}\sum_{\substack{p, q \in P_i \\ p \neq q}} f_\theta (p, q) \\
    \end{aligned}
\end{equation}

Imagine a situation where we are deciding between two clusters, $P_1$ and $P_2$, all elements in both $P_1$ and $P_2$ are very close to $v$ according to $f_\theta$, and $P_2$ contains an outlier. If we simply weigh down the influence of the outlier, the likelihoods for $P_1$ and $P_2$ would be roughly equivalent, but the mere fact that $P_2$ even contains an outlier means that the patient's data is inconsistent or mislabeled. In any case, we lose confidence that assigning $v$ to $P_2$ is the right decision. Therefore, in order to decide to add $v$ to $P_2$, the values of $f_\theta$ between $v$ and other cluster elements should be large enough to outweigh this.

Note that normalizing $c_i$ itself inside one cluster is redundant, as any positive (not sign-changing) linear transformation of $l(v, P)$ is monotonic and therefore does not influence the value of its argument maxima. $c_i$ simply has to be normalized by any scalar multiple of $\frac{1}{n(n-1)}$ so that it is normalized with respect to the size of $P_i$, as different patients can have different amounts of ECG vectors.


\section{Experiments}

For training of our model in both phases, we use the largest dataset available to us -- the CODE-15\%. Given the fact that the PTB and PTB-XL datasets were collected by the same institute, it is reasonable to assume higher dependence between their samples. Hence, we have opted out of including any part of PTB in the validation set. Since we aim to produce a model that generalizes well, we never train it on data outside of CODE-15\% and use that data for verification of its ability to generalize. If we had trained the model using a combination of CODE-15\% and other large datasets, we might have obtained better results. However, doing so could have undermined our confidence in the validity of the results, as it would have been difficult to ascertain whether the model wasn't simply adapting to the unique features and systematic noise of those specific three datasets.


The amount of both possible A/P/N \textit{triplets} and \textit{pairs} is (due to the ratio between patients and signals) approximately quadratic in terms of the number of patients, making its size impractically large for enumeration. Hence, during training and validation, we \textit{sample} the sets of triplets and pairs instead of enumerating them sequentially. Sampling of positive (same-patient) and negative (different-patients) pairs is strictly balanced. To ensure deterministic validation, we seed the triplet and pair generators with a fixed value and early-stop the training procedure on the validation loss of CODE-15\%. The specific part of the dataset that we sample from is defined by the train/dev/test split.

\subsection{Optimal configuration}

Since our goal is to find out which configuration generalizes the best, we tune various model hyper-parameters and look for the combination that maximizes the \textit{minimal} AUC measure across all three large datasets. The results of this hyper-parameter search are captured by Tables \ref{tab:hpsearch-1},  \ref{tab:gridsearch-1} and \ref{tab:gridsearch-2}. Due to the vast range of possible combinations, we have first identified an optimal configuration for the embedding model using a full discriminator without a hidden layer, and then optimized the configuration for the discriminator separately.

\begin{landscape}

\begin{table}
    \centering
    \begin{tabular}{lllll}
        \toprule
        \textbf{Hyper-parameter} & \textbf{Options} & \textbf{Best} & \textbf{$\Delta_{AUC}$} & \textbf{p-value}\\
        \midrule
        Model & 1D-RN, CDIL & CDIL & 0.013 & 1e-8 \\ 
        Embedding loss (EL) & triplet, circle & triplet & 0.016 & 3e-6 \\
        Embedding size (ES) & 128, 256, 384, 512 & 256 & 0.002 & 0.151 \\ 
        Embedding fine-tuning (EF) & freeze, end-to-end & end-to-end & 0.004 & 0.037 \\
        Normalization (NORM) & apply, exclude & apply & 0.031 & 3e-8\\ 
        Baseline wander removal (BWR) & apply, exclude & exclude & 0.026 & 4e-10\\
        High-freq. noise removal (HFNR) & apply, exclude & apply & 0.001 & 0.322\\
        \midrule
        Discriminator hidden size (DHS) & 0 (exclude), 16 & 16 & 0.007 & 1e-3\\
        Discr. $l_1$ distance (DL1) & exclude, merge, full & full & 0.412 & 2e-7\\
        Discr. $l_2$ distance (DL2) & exclude, merge, full & exclude & 0.002 & 0.044 \\
        Discr. $cos$ distance (DCOS) & exclude, merge, full & exclude & 0.001 & 0.086 \\
        \bottomrule
    \end{tabular}
    \caption{All examined hyper-parameters and their options. Embedding fine-tuning represents a setting where, in the Siamese model, the embedding model is either frozen or allowed to be trained end-to-end during the second phase of training. The last column, $\Delta_{AUC}$, is the difference in performance between the best configuration and the best alternate configuration across all other options for each specific hyper-parameter. For example, the best configuration's performance average is better by 0.031 (AUC) than the same configuration but without normalization. The p-value is that of a one-tailed t test between those corresponding hyper-parameter setting experiments, where $H_A$ is that the best configuration is better than the alternative one. All models were trained using the Adam optimizer with learning rate 0.001, without any LR scheduler or weight regularization.}
    \label{tab:hpsearch-1}
\end{table}

\end{landscape}

\begin{table}[!htb]
    \footnotesize	
    \begin{subtable}{\linewidth}
      \centering
        \begin{tabular}{lllllllrr}
            \toprule
            \textbf{Model} & \textbf{EL} & \textbf{ES} & \textbf{NORM} & \textbf{BRW} & \textbf{HFNR} & \textbf{EF} & \textbf{AUC} & \textbf{SEM} \\
            \midrule
            CDIL &  triplet &            128 &      apply &                   exclude &                   apply &             end-to-end &  0.9496 &  0.0010 \\
            CDIL &  triplet &            384 &      apply &                   exclude &                  exclude &              freeze &  0.9450 &  0.0012 \\
            CDIL &  triplet &            128 &      apply &                   exclude &                  exclude &             end-to-end &  0.9501 &  0.0016 \\
            CDIL &  triplet &            256 &      apply &                   exclude &                  exclude &             end-to-end &  0.9507 &  0.0011 \\
            CDIL &  triplet &            256 &      apply &                   exclude &                   apply &             end-to-end &  0.9516 &  0.0015 \\
            \bottomrule
        \end{tabular}
        \caption{Top 5 embedding model hyper-parameter grid-search results. }
        \label{tab:gridsearch-1}
    \end{subtable}

    \bigskip
    
    \begin{subtable}{\linewidth}
      \centering
        \begin{tabular}{llllrr}
            \toprule
            \textbf{DCOS} & \textbf{DHS} & \textbf{DL1} & \textbf{DL2} &  \textbf{AUC} & \textbf{SEM} \\
            \midrule
            merge &                        16 &                      full &                     merge &  0.9588 &  0.0016 \\
            exclude &                        16 &                      exclude &                      full &  0.9593 &  0.0011 \\
            merge &                        16 &                      full &                      exclude &  0.9594 &  0.0017\\
            full &                        16 &                      full &                      exclude &  0.9595 &  0.0005\\
            exclude &                        16 &                      full &                      exclude &  0.9612 &  0.0010\\
        \bottomrule
        \end{tabular}
        \caption{Top 5 discriminator head hyper-parameter grid-search results.}
        \label{tab:gridsearch-2}
    \end{subtable} 
    \caption{The best 5 configurations of both models. AUC is the minimal AUC across all three large datasets. SEM stands for Standard Error of the Mean $= \frac{\sigma}{\sqrt{n}}$, which can be used for statistical testing. All discriminator head settings use the best embedder model hyper-parameters. The full table can be found at our GitHub repository.}
\end{table}

Regarding the embedding model, we found that the filtering methods we have experimented with have not been beneficial to the model's performance, especially the baseline wander removal filter, which actually hurts the performance ($-0.03$ AUC). The CDIL-CNN architecture is already resistant to noise, and thus the filters are largely redundant and may even corrupt the original signal in special occasions (for example when iterative filters, such as our wavelet-based BWR, fail to converge). 

It is also impossible to say whether freezing the embedding model after the first stage helps or not; it would certainly be beneficial to freeze it if overall training times were a major factor. We have trained our models on various GPU cards in the Metacentrum cloud computing grid service and the overall training time never surpassed 18 hours. The same goes for the model's embedding size, which has turned out to be an insignificant parameter in our setting. We have settled on 256, but reducing the size to 128 should have negligible impact on the system's performance while reducing the required storage space for patient vectors by half.

Note the large value of $\Delta_{AUC}$ for the discriminator $l_1$ distance parameter. It shows that not letting the model adjust the distance member weights -- that is, literally computing the distance between the input ECG vectors -- significantly hurts its performance. We can also see that it is likely not optimal to incorporate further distance metrics into the discriminator head, as that causes the model to over-fit slightly and worsen its generalization capabilities. However, this claim is not backed by statistical testing, as we fail to reject the hypothesis that the best setting should exclude the cosine distance completely under confidence level 95\%. But even under the assumption that the discriminator indeed should use only a single distance measure, in Table \ref{tab:gridsearch-2}, we can see that $l_1$ is not clearly superior to $l_2$ in this sense.

In an attempt to further regularize the model and thus cause it to generalize better, we have shortly experimented with shuffling the electrocardiogram leads and adding a small amount of Gaussian noise to them. However, these techniques have led to a significant decrease of performance (-0.02 to -0.06 AUC) and have thus been abandoned.



\subsection{Evaluation}

Finally, we measure the model's performance on the two evaluation tasks: gallery-probe matching, and overseer simulation. The results are shown in Table \ref{tab:final-results}. 

For gallery-probe matching, we have selected a random sample of patients with at least two recordings from the test set of each respective dataset. The exception is PTB-XL, where the gallery and probe sets were built from the whole dataset like in the study by Oh et al. \cite{oh2022lead} for the purpose of meaningful and fair comparison. It should be noted that although the optimal model configuration was selected based on the minimum dev set performance across all three large datasets, the minimum performance was never that of PTB-XL, which allows us to perform gallery-probe evaluation on the entirety of PTB-XL without compromising the validity of our results. The sample size is included in the aforementioned table. 

For overseer simulation, we select 10000 patients to initialize the database with and 1000 further electrocardiograms to classify (probe) under staff mistake rate $2\%$, meaning that there are 20 mistakes to detect among 980 correct classifications. The mistake rate of 2\% is our best estimate of the real mistake rate occurring in practice, as it is the difference in pairwise accuracy of our model on PTB-XL and IKEM. Due to the size requirements, PTB was excluded from the experiments, as it contains only 290 patients, and the number of initial patients for PTB-XL was only $\approx 9400$. 

Since we find out that weighting the discriminator outputs by representational quality (\aC) performs the best (Table \ref{tab:approach-comparison}), we consequently report results that were obtained under this approach. We hypothesize that the drop of performance in \aD\ compared to \aC\ is caused by this approach being biased towards clusters of size 1 (where consistency is maximized) regardless of whether they are a good fit for the classified ECG vectors or not.

\begin{table}[h]
    \centering
    \begin{tabular}{ll}
        \toprule
        \textbf{Approach} & \textbf{P@R95 (median)} \\
        \midrule
        \aA & 3.66\% \\
        \aB & 18.35\% \\
        \aC & \textbf{22.35\%} \\
        \aD & 7.14\% \\
        \bottomrule
    \end{tabular}
    \caption{Comparison of likelihood calculation approaches in the overseer simulation task aggregated across all datasets (CODE-15\%, IKEM, PTB-XL) and mistake rates (50\%, 5\%, 2\%, 1\%). }
    \label{tab:approach-comparison}
\end{table}

\begin{table}[h]
    \footnotesize
    \centering
    \begin{tabular}{llllll}
        \toprule
        \textbf{Task} & \textbf{Metric} & \textbf{Dataset} & \textbf{} & \textbf{} & \textbf{}\\ \\
        \textbf{} & \textbf{} & {CODE-15\%} & {IKEM} & {PTB-XL} & {PTB}\\
        \midrule \midrule
        training objective & {AUROC} & 0.990 & 0.971 & 0.984 & 0.982 \\
        \textit{} & {Accuracy} & 95.8\% & 92.1\% & 94.3\% & 93.1\% \\
        \midrule
        gallery-probe & {Accuracy} & 60.3\% & 46.0\% & 58.3\% & 77.0\% \\
        & Sample size & 2127 & 2127 & 2111 & 113 \\
        \midrule
        overseer simulation & {P@R95} & $0.28 \pm 0.09$ & $0.08 \pm 0.02$ & $0.17 \pm 0.08$ & \multicolumn{1}{c}{--} \\
        \textit{} & {F1} & $0.59 \pm 0.04$ & $0.39 \pm 0.06$ & $0.56 \pm 0.06$ & \multicolumn{1}{c}{--} \\
        \textit{} & {CR} & $45\% \pm 6\%$ & $7\% \pm 3\%$ & $19\% \pm 5\%$ & \multicolumn{1}{c}{--} \\
        \bottomrule
    \end{tabular}
    \caption{Overview of the results obtained on test sets across all datasets and evaluation tasks, including the original training objective (pairwise matching). In the context of pairwise matching, the decision threshold is optimized for maximizing accuracy on the dev set. For overseer simulation, we report 95\% confidence intervals of the mistake detection rates, where P@R95 stands for precision at recall 95, F1 is the F-measure between precision and recall under threshold obtained by achieving recall 95 on the dev set, and CR is the fraction of corrected mistakes (the patient with highest likelihood was the real owner) out of all detected mistakes.}
    \label{tab:final-results}
\end{table}

\begin{table}[h]
    \footnotesize
    \centering
    \begin{tabular}{l|ll|ll|ll}
        \toprule
        {Dataset} & \multicolumn{2}{c}{CODE-15\%} & \multicolumn{2}{c}{IKEM} & \multicolumn{2}{c}{PTB-XL}\\
        \midrule \midrule
        \backslashbox{{Gold}}{{Pred}} & {{true}} & {{false}} & {{true}} & {{false}} & {{true}} & {{false}}\\
        \midrule
        {true} & 18 & 2 & 5 & 15 & 9 & 11 \\
        {false} & 22 & 958 & 2 & 978 & 3 & 977 \\
        \bottomrule
    \end{tabular}
    \caption{Confusion matrices of our model in overseer simulation on the test set parts of each dataset. Gold truth value in this context signifies whether the simulated clinician made a mistake and predicted truth value signifies whether the overseer model reported a mistake.}
    \label{tab:om-conf-matrix}
\end{table}

\begin{table}[h]
    \footnotesize
    \centering
    \begin{tabular}{lll}
        \toprule
        \textbf{Method} & \textbf{Author(s)} & \textbf{Accuracy} \\
        \midrule \midrule
        random guessing & - & $\le$ 0.05\% \\
        \midrule 
        SimCLR & \textit{Chen et al. (2020)} \cite{chen2020simple} & 35.3\% $\pm$ 0.3\% \\
        3KG & \textit{Gopal et al. (2021)} \cite{gopal20213kg} & 40.7\% $\pm$ 0.4\% \\
        Wav2Vec 2.0 (W2V) & \textit{Baevski et al. (2020)} \cite{baevski2020wav2vec} & 49.2\% $\pm$ 0.4\% \\
        CMSC & \textit{Kiyasseh et al. (2021)} \cite{kiyasseh2021clocs} & 51.3\% $\pm$ 0.6\% \\
        W2V + CMSC + RLM & \textit{Oh et al. (2022)} \cite{oh2022lead} & 57.7\% $\pm$ 0.6\% \\
        \midrule
        ElectroCardioGuard & \textit{ours} & \textbf{58.3\% $\pm$ 0.6\%} \\
        \bottomrule
    \end{tabular}
    \caption{Comparison of our model to the state of the art in PTB-XL gallery-probe patient ECG matching. Here, CMSC and RLM are techniques applied by Oh et al. \cite{oh2022lead} in order to enhance the accuracy of their model, which are sampling positive pairs from the same ECG recording (CMSC \cite{kiyasseh2021clocs}) and randomly masking some of the 12 ECG leads (RLM). The \textit{Accuracy} values measure the probability that any probe element is correctly matched to its gallery element by the corresponding model or technique, with 95\% confidence interval.}
    \label{tab:gallery-probe-comparison}
\end{table}

Notably, we can see that the evaluation tasks are much harder than the training objective due to the vast amount of ECG vector pairs the model must correctly recognize as true negatives. During pairwise training and evaluation of the discriminator, positive (same patient) and negative pairs (different patients) are sampled in a balanced manner. However, in the gallery-probe matching task, the discriminator must assign a probability value to a single positive pair that is higher than those of thousands of other negative pairs, which are created between the probe element and the entire gallery. The situation is similar in the overseer simulation task except that there are a few more positive pairs.

Our efforts to minimize the difference in accuracy in the pairwise task across all datasets have resulted in a model with the highest degree of generalization across all three datasets. However, we can notice that even a small decrease in pairwise accuracy has a significant impact on performance in the evaluation tasks, which are much harder. It follows that, for optimal performance in production, institutions should invest in fine-tuning our model on their specific datasets using our published version as a robust starting point instead. In such cases, according to Table \ref{tab:final-results}, if we wanted to detect 19 out of 20 mistakes made in 1000 classifications, our model would cause only 1 false alarm in roughly 23 classifications, which the hospital staff would have to dismiss. \todo{Added the following sentence (Given the fact...), including the referenced new Table \ref{tab:om-conf-matrix}} Given the fact that we have set the model's decision threshold in order to achieve the same goal (19 out of 20 mistakes caught) using the dev set before finally measuring its performance on the test set, then according to Table \ref{tab:om-conf-matrix}, we caught only 18 out of 20 mistakes but caused only 1 false alarm in roughly 45 classifications instead of 23.

Our model's capacity for generalization is also demonstrated by its performance in gallery-probe matching and overseer simulation (F1) on PTB-XL. Despite never encountering a single example from PTB-XL, our model achieved comparable performance to CODE-15\%, which encompasses the entire training dataset. The significant performance drop observed in IKEM could be caused by its different patient-to-recording ratio (roughly twice as much recordings per patient on average, see Table \ref{tab:datasets}) or noisier labels.

Furthermore, our model's performance in the gallery-probe task on PTB-XL matches the state of the art set by Oh et al. \cite{oh2022lead} (12 leads, best variant), but using a significantly smaller number of parameters. It is difficult to specify precisely how much smaller our model is without knowing the exact number of parameters of the state of the art model, but seeing that it contains BERT-base (110M parameters) without the embedding layer (24M parameters) and includes additional convolutional layers, we can determine that our model is at least 760 times smaller, if not more. \todo{Added a reference to the new Table \ref{tab:gallery-probe-comparison} to the next sentence (As shown in...)} As shown in Table \ref{tab:gallery-probe-comparison}, our model even surpasses the state of the art results by 0.6\%, however, the improvement is not statistically significant. Furthermore, this small improvement could be attributed to the fact that the current version of PTB-XL (1.0.3) contains 16 less patients with at least two recordings than the older version used by Oh et al.

\todo{Added the following paragraph (All our training...)}
All our training experiments were executed on a Metacentrum (MetaVO) cluster using Tesla T4 14,9 GiB GPU cards, using batch size 128 or 256, and all runs finished within 8 hours. However, regarding the \textit{deployment} of our automated ECG misclassification detection system, there are two important factors to consider. Firstly, the computational cost of running our deep learning model should be taken into account, especially when dealing with a large number of ECG recordings. In the context of our overseer simulation, which builds a database of patients and simulates a clinician assigning new ECGs to patients, hence resembling the intended real-life use case, we have conducted a CPU benchmarking test. The chip we used for this purpose was an Intel(R) Core(TM) i5-4440 CPU @ 3.10GHz\footnote{At the time of writing this paper, this chip is considered below median in terms of processing power \cite{cpu-benchmark}.} and we found that a single check whether a clinician makes a mistake or not (consisting of multiple forward passes) averages at 0.4ms without optimization. Given our model's modest size and the benchmarked efficiency, our system is well-suited for centralized or local on-premise CPU inference, without the need for extensive investments in memory, GPUs, or cloud solutions. Secondly, this approach addresses privacy concerns that arise when processing sensitive medical data. By maintaining all data within the secure confines of the medical facility's storage, the suggested deployment strategy mitigates the risk of private data exposure.

\section{Discussion}

In summary, our work has contributed to the field of automatic electrocardiogram processing and patient identification in multiple ways. 

First, we publish a large brand-new electrocardiogram patient identification dataset with many recordings per patient on average, which is suitable for training and evaluation of models based on electrocardiogram representations. The dataset is located in the \verb|datasets| folder of our public GitHub project.

Second, we publish a tiny (700 kB) neural-network model utilizing state-of-the-art techniques for sequence processing capable of deciding whether two ECG recordings originate from the same individual. We further show how can our model be used for misclassification detection by employing it in a streamed clustering process, and evaluate it in a simulation mimicking the real application. Both the simulation and various clustering approach implementations are published as a stand-alone part of our GitHub project. This was the primary goal of our study, which was primarily motivated by the Institute for Clinical and Experimental Medicine's sponsorship, as they are actively preparing to implement our model in their production environment.

Our model demonstrates applicability not only in the intended scenario but also in a variety of other cases. For instance, our model can be employed to address the issue of fixing existing databases, rectifying inconsistencies and improving data quality. Moreover, it provides a solution for situations where a patient's record is missing within a specific time frame, allowing us to identify the inconsistency by cross-referencing the date of addition and determining which 100 patients were present on that day. Another valuable application is the ability to detect potential patient swaps that may occur on a particular day, which is a task similar to gallery-probe matching, but much easier in difficulty. This feature can be especially useful in quickly identifying and resolving any mix-ups, thanks to the capabilities of our system. Lastly, our model can address challenges encountered in remote areas where doctors may not have access to their own electrocardiogram machines. By verifying the integrity of electrocardiogram measurements obtained through external requests, it helps mitigate the risk of local mixing of recordings.

\todo{Added subsection Limitations}
\subsection{Limitations}
However, our method has its drawbacks. Our further experiments show that our model struggles to correctly distinguish between patients if the ECG sampling frequency becomes too low or too high; 350-750 Hz being the optimal range. If our model were to be applied in such context, we recommend either re-sampling the ECG readings to 500 Hz before inputting them to our network, or fine-tuning the network on a small portion of the relevant data. The latter may be the optimal approach, because although we have shown that our model generalizes well beyond the scope of the domain of our training dataset, its performance is not optimal there. Otherwise, our model is resistant to noisy ECGs containing various artifacts, which it has been trained to process and filter out.

In our study, we did not experiment with the influence that various cardiovascular conditions may have on the performance of our model. This was primarily caused by the specific format of the IKEM dataset at the time of publishing of our paper (ECG recording and patient ID), which provides no additional information regarding the patient's health. In the near future, we plan to co-operate with the researchers at the Institute of Clinical and Experimental Medicine to obtain health-related annotations for the IKEM dataset and conduct a follow-up study.

\todo{Added subsection Ethical issues}
\subsection{Ethical issues}
The application of any patient identification method poses several ethical challenges, such as its reliability, veracity, and biases or discrimination of certain ethical groups, which are problems commonly occurring in machine-learning solutions making use of human-labeled data. However, due to the fact that none of our data includes any human labels, we feel confident that our model will not exhibit biased behavior. Furthermore, our model is intended to be used only as a tool which attempts to detect human classification errors. As such, it is not directly responsible for diagnosing patients with heart conditions or classifying patients itself, but serves merely as an indicator of possible mistakes that the clinicians may have caused, giving them chance to correct themselves. Any false positives reported by the model can be easily and inconsequentially dismissed by the clinicians, whilst its false negatives are errors that would have been missed regardless. In practice, we advise setting our model up in a manner that prioritizes the reduction of false negatives over false positives, mirroring the approach taken in our overseer simulation. This strategic choice can lead to cleaner ECG databases, ultimately enhancing the quality of patient care.

\section*{Acknowledgement}
\noindent Computational resources were supplied by the project "e-Infrastruktura CZ" (e-INFRA LM2018140) provided within the program Projects of Large Research, Development and Innovations Infrastructures.

\vspace{0.9em}

\noindent This work has been supported by Grant No. SGS-2022-016 Advanced methods of data processing and analysis.

\section*{Conflicts of interest}

\noindent The authors declare the following potential conflicts of interest:

Michal Seják and David Žahour received funding from the Institute for Clinical and Experimental Medicine (IKEM) to conduct this research project on detection of patient misclassifications using electrocardiogram recordings.
IKEM provided financial support for the research and covered data collection. The funding from IKEM did not involve any restrictions on study design, data analysis, or result interpretation.

It is important to note that despite the funding received from IKEM, the research was conducted independently and the authors maintained full control over the study design, data analysis, and decision to publish. The authors affirm that the study was conducted with scientific rigor, objectivity, and integrity, adhering to established research protocols and methodologies. The evaluation of the model was performed using publicly available datasets, ensuring transparency and minimizing potential biases.

We would like to acknowledge the support provided by IKEM and express our gratitude for their financial assistance in conducting this research. However, the funders had no role in the study design, data analysis or pre-processing, manuscript preparation, or decision to submit for publication.

Please note that all authors have reviewed and approved the contents of this disclosure.

\bibliographystyle{elsarticle-num} 
\bibliography{main}
 
\clearpage
\appendix

\section{Appendix}

\begin{figure}[h]
    \centering
    \includegraphics[width=.7\textwidth]{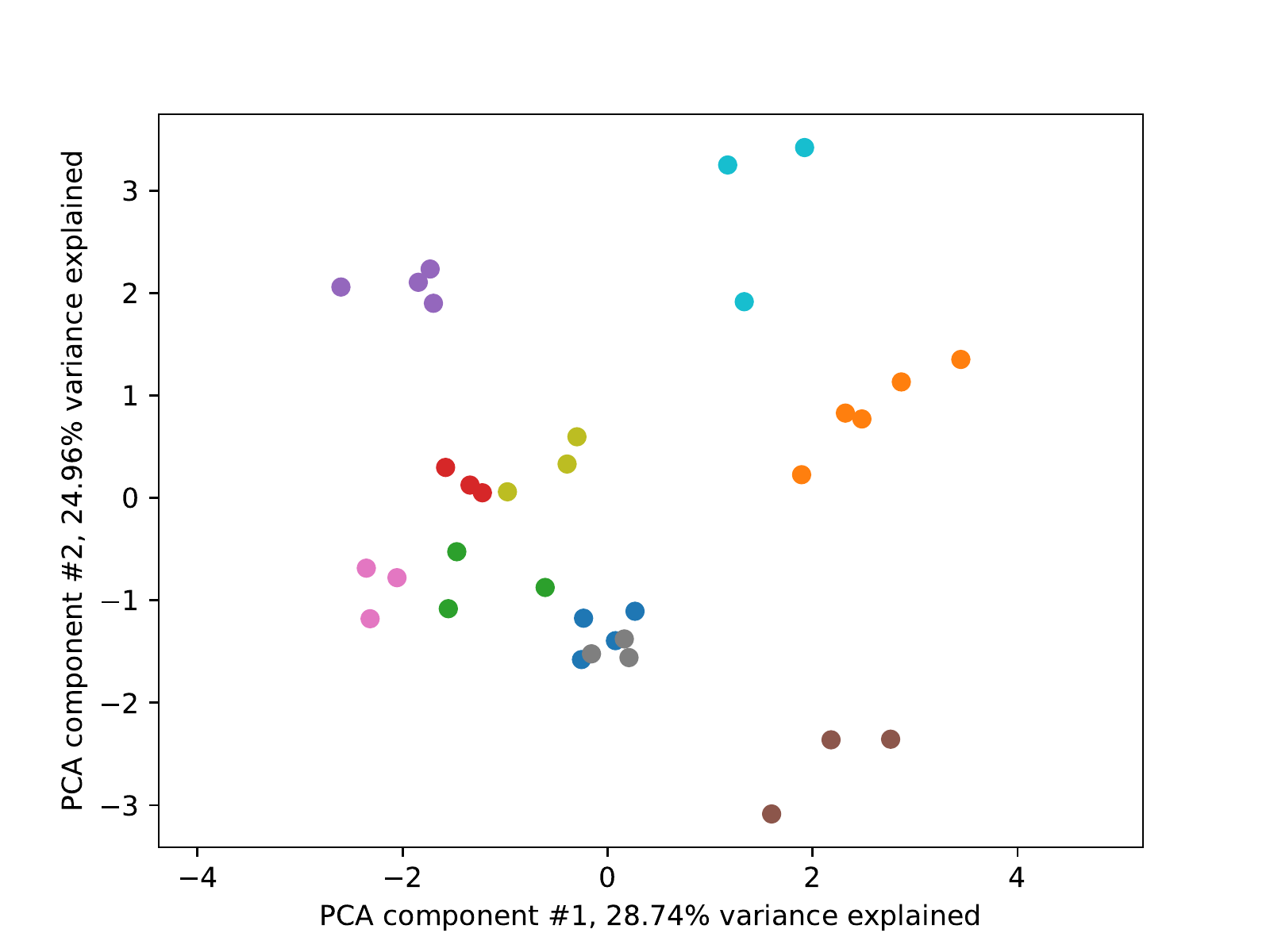}
    \caption{10 random patients with at least 3 recordings from PTB-XL embedded using our embedding model and reduced to two dimensions using PCA.}
    \label{fig:pca-embeddings}
\end{figure}

\begin{figure}[h]
    \centering
    \begin{subfigure}[b]{0.45\textwidth}
    \centering
    \includegraphics[width=\textwidth]{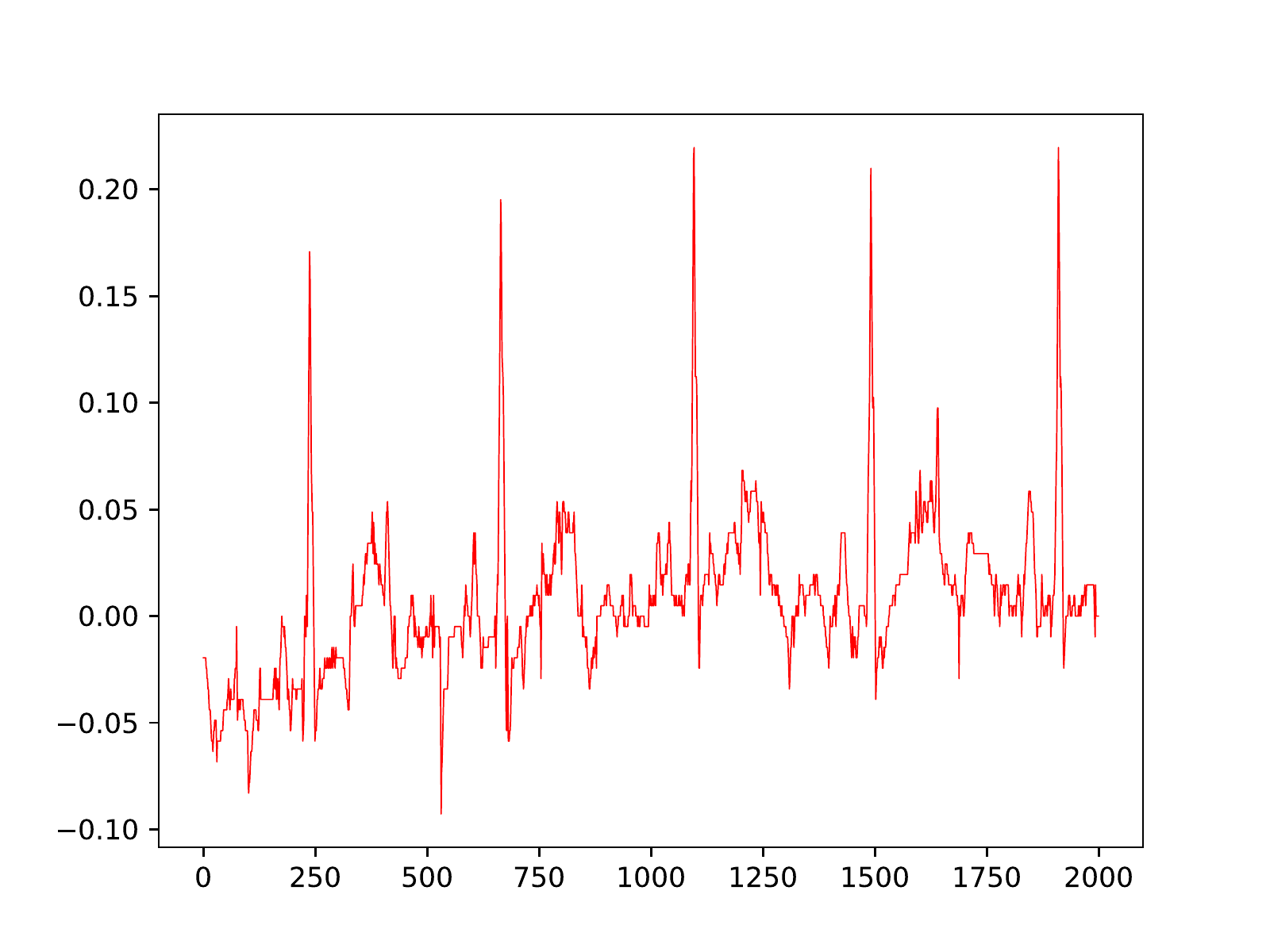}
    \caption{Raw ECG signal (example from PTB-XL).}
    \vspace{2.1em}
    \label{fig:pp1}
    \end{subfigure}
    \hfill
    \begin{subfigure}[b]{0.45\textwidth}
    \centering
    \includegraphics[width=\textwidth]{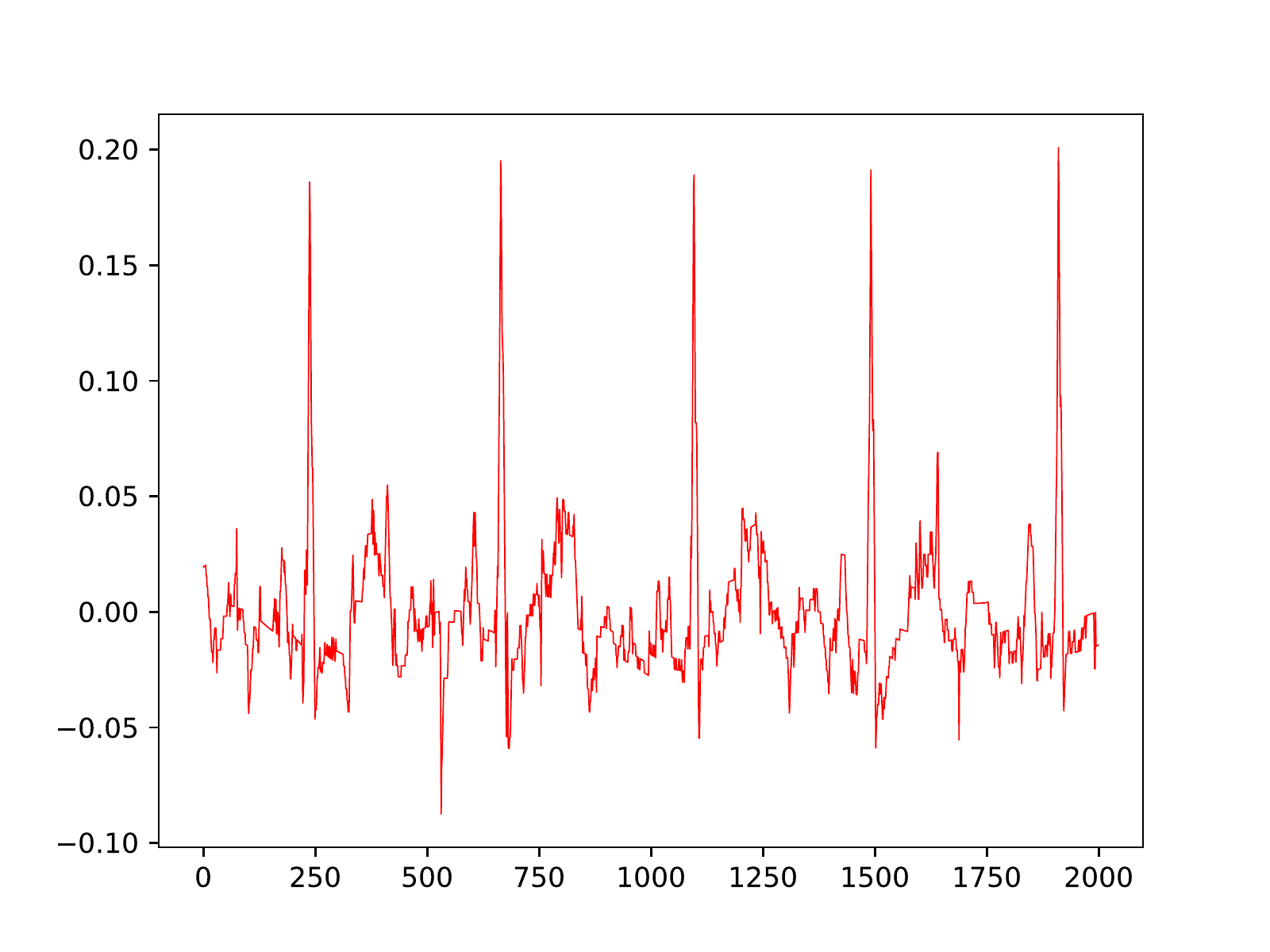}
    \caption{Baseline wander removal. The resulting signal's baseline is now a straight line $y=0$ as opposed to the raw signal, where it starts at $-0.05\ \mu V$ and gradually rises. }
    \label{fig:pp2}
    \end{subfigure}    
    \begin{subfigure}[b]{0.45\textwidth}
    \centering
    \includegraphics[width=\textwidth]{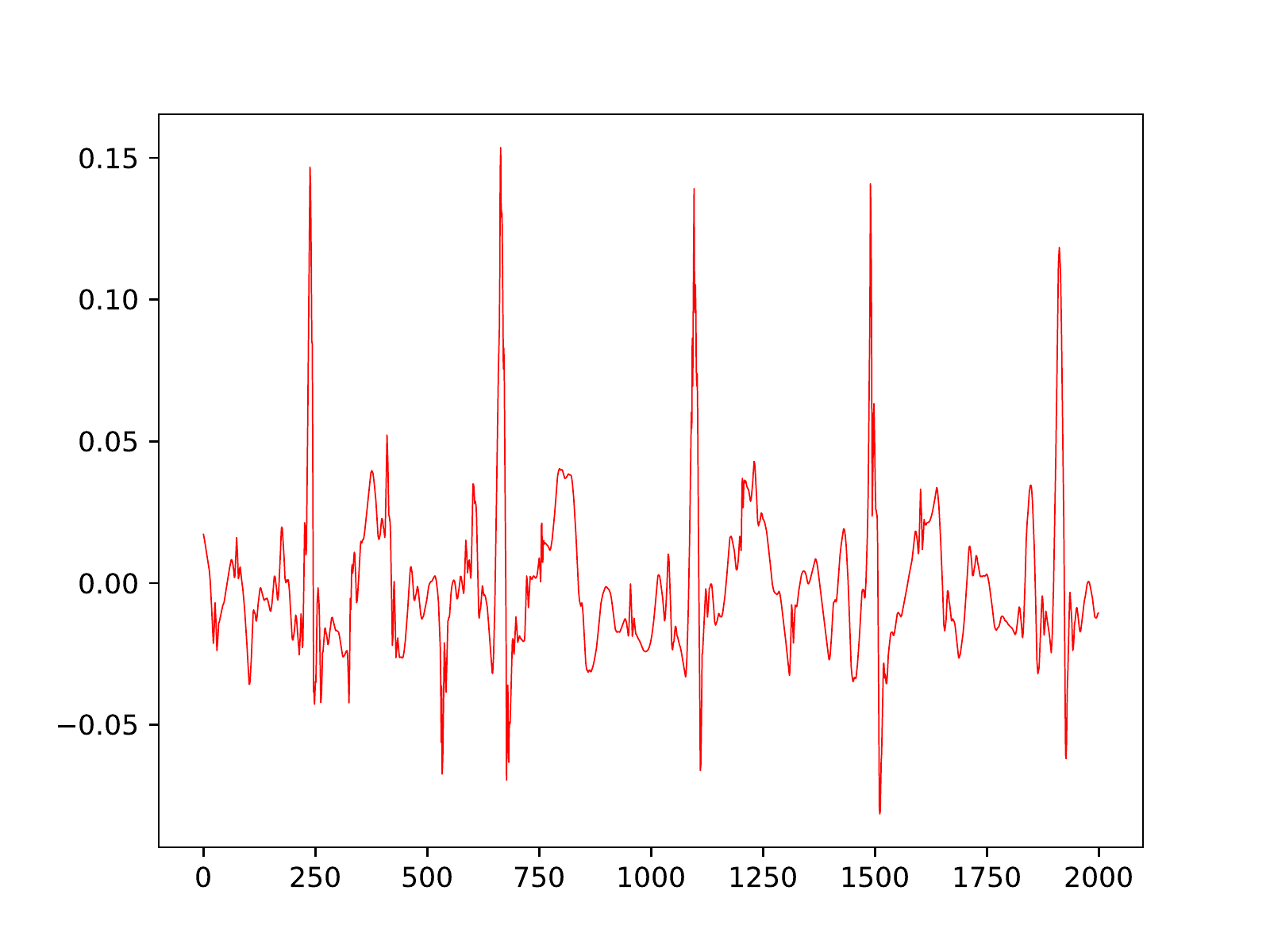}
    \caption{High-frequency filtering. Notice how the small high-frequency oscillations in the original signal are now gone and the resulting signal curve is smoother. }
    \label{fig:pp3}
    \end{subfigure}
    \hfill
    \begin{subfigure}[b]{0.45\textwidth}
    \centering
    \includegraphics[width=\textwidth]{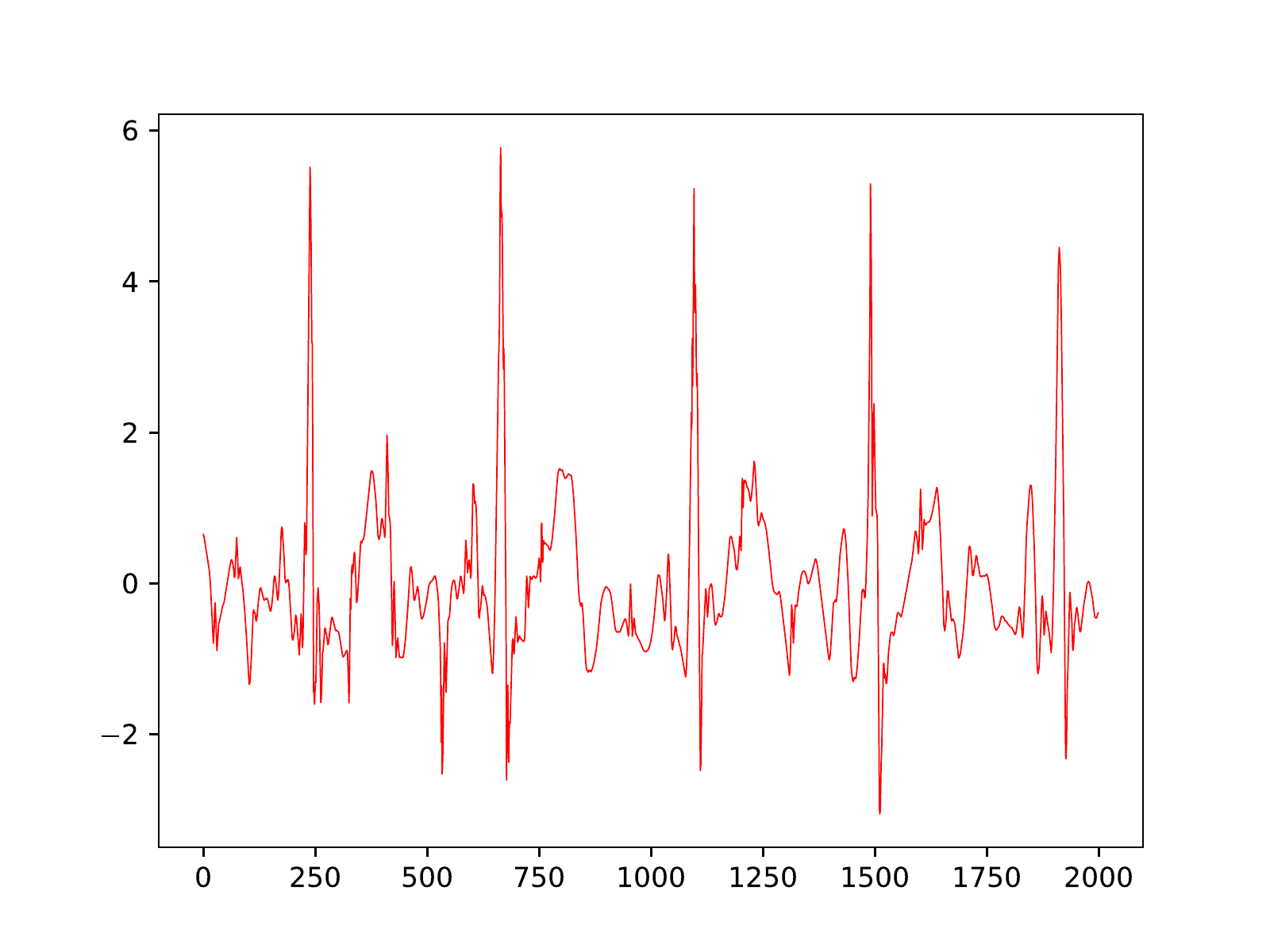}
    \caption{Normalization. The Y-axis ticks are re-calibrated to count the number of standard deviations from the signal's mean.}
    \vspace{0.9em}
    \label{fig:pp4}
    \end{subfigure}
    \caption{The pre-processing methods we have examined. In our setting, they are always chained in the above order, where the initial signal is taken from the dataset and the final result is input to our embedding model.}
    \label{fig:preprocessing}
\end{figure}

\end{document}